\documentclass{IEEEtran4PSCC}
\pdfoutput = 1

\ifCLASSINFOpdf
   \usepackage[pdftex]{graphicx}
\else
   \usepackage[dvips]{graphicx}
\fi
\usepackage[cmex10]{amsmath}

\usepackage{breqn}

\usepackage{amsfonts}
\usepackage{subfig}
\usepackage{comment}
\usepackage[table,xcdraw]{xcolor}

\makeatletter
\let\old@ps@headings\ps@headings
\let\old@ps@IEEEtitlepagestyle\ps@IEEEtitlepagestyle
\def\psccfooter#1{%
    \def\ps@headings{%
        \old@ps@headings%
        \def\@oddfoot{\strut\hfill#1\hfill\strut}%
        \def\@evenfoot{\strut\hfill#1\hfill\strut}%
    }%
    \def\ps@IEEEtitlepagestyle{%
        \old@ps@IEEEtitlepagestyle%
        \def\@oddfoot{\strut\hfill#1\hfill\strut}%
        \def\@evenfoot{\strut\hfill#1\hfill\strut}%
    }%
    \ps@headings%
}
\makeatother

\psccfooter{%
        \parbox{\textwidth}{\hrulefill \\ \small{23rd Power Systems Computation Conference} \hfill \begin{minipage}{0.2\textwidth}\centering \vspace*{4pt} \includegraphics[scale=0.06]{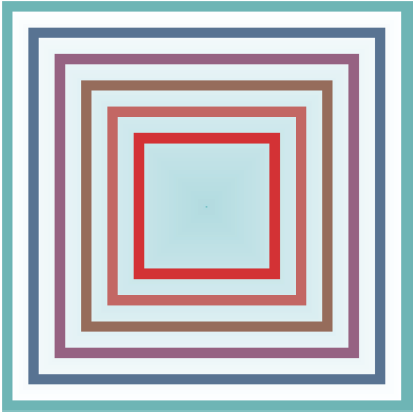}\\\small{PSCC 2024} \end{minipage} \hfill \small{Paris, France --- June 4 -- 7, 2024}}%
}

\begin{document}
%
 \title{ Conformal Prediction for Stochastic Decision-Making of PV Power in Electricity Markets}

\author{\IEEEauthorblockN{Yvet Renkema\IEEEauthorrefmark{1}, Nico Brinkel\IEEEauthorrefmark{2} \& Tarek Alskaif\IEEEauthorrefmark{1}\\ \IEEEauthorrefmark{1}Information Technology Group, Wageningen University, The Netherlands,\\
\IEEEauthorrefmark{2}Copernicus Institute of Sustainable Development, Utrecht University, The Netherlands
}}

\maketitle

\begin{abstract}
This paper studies the use of conformal prediction (CP), an emerging probabilistic forecasting method, for day-ahead photovoltaic power predictions to enhance participation in electricity markets. First, machine learning models are used to construct point predictions. Thereafter, several variants of CP are implemented to quantify the uncertainty of those predictions by creating CP intervals and cumulative distribution functions. Optimal quantity bids for the electricity market are estimated using several bidding strategies under uncertainty, namely: trust-the-forecast, worst-case, Newsvendor and expected utility maximization (EUM). Results show that CP in combination with k-nearest neighbors and/or Mondrian binning outperforms its corresponding linear quantile regressors. Using CP in combination with certain bidding strategies can yield high profit with minimal energy imbalance. In concrete, using conformal predictive systems with k-nearest neighbors and Mondrian binning after random forest regression yields the best profit and imbalance regardless of the decision-making strategy. Combining this uncertainty quantification method with the EUM strategy with conditional value at risk (CVaR) can yield up to 93\% of the potential profit with minimal energy imbalance.
\end{abstract}

\begin{IEEEkeywords}
Conformal prediction; Electricity markets; Machine learning; Photovoltaic power; Stochastic optimization
\end{IEEEkeywords}

\thanksto{\noindent Submitted to the 23rd Power Systems Computation Conference (PSCC 2024).}

\section{Introduction}
In the day-ahead market (DAM) for electricity, suppliers offer a specific volume of electricity that they will be able to supply the next day along with their minimum selling price. Meanwhile, buyers bid a volume that they are willing to receive the following day along with a maximum buying price. In the Netherlands, this auction-based market mechanism has an hourly resolution and closes at noon the day before delivery \cite{epexspot}. If suppliers or buyers deviate from their accepted bid in real-time, they are subject to high imbalance costs in the real-time market (RTM). Therefore, it is generally beneficial for electricity suppliers to avoid deficiencies and surpluses. For this reason, reliable solar photovoltaic (PV) power forecasts are required for electricity suppliers with PV assets. Similarly, grid operators require insight into the expected PV generation to make estimates about the congestion levels in their network. 

The rapidly increasing adoption of PV increases the overall uncertainty in power markets, underscoring the necessity for accurate PV power predictions. Extensive research, as highlighted in literature reviews \cite{LitRevForecastingML, LitReviewPVForecasting}, has focused on solar irradiance and PV power forecasting. Probabilistic forecasting enhances the reliability of PV power predictions by providing information about their full probability distributions \cite{van2018review}. This is particularly important for decision-making in electricity markets where uncertainty plays a significant role \cite{morales2013integrating}. 

Conformal prediction (CP) is an emerging distribution-free and model-agnostic probabilistic forecasting method that offers a measure of confidence by transforming point predictions into prediction intervals \cite{GentleIntro}. In its most basic form, the residuals of predictions from a calibration dataset are used to calibrate prediction intervals from a test dataset so those intervals have a probabilistic guarantee of covering the true outcome. This makes CP particularly useful in applications with uncertainty where reliability is essential, such as decision-making on electricity markets. Various adaptations of CP contribute to conditional guarantees, making it a very flexible method. The application of CP for decision making under uncertainty is rather new. One study utilized  CP to minimize a worst-case scenario based on simulated data \cite{johnstone2021conformal}. Furthermore, CP has been employed to optimize prediction intervals for gas demand, albeit without connecting this to decision-making in the energy market \cite{pmlr-v179-mendil22a}. To the authors' knowledge, there is no study utilizing CP for PV power predictions or directly linking the quantified uncertainty of those predictions to decision-making in electricity markets.

This paper proposes and investigates the added value of a framework for stochastic decision-making considering PV power participation in electricity markets. This framework is based on the implementation of point prediction models that predict PV power day-ahead based on weather forecasts. This is then combined with CP to quantify the uncertainty of point predictions in the form of prediction intervals or cumulative distribution functions (CDF). 
Subsequently, several bidding strategies, including Trust-the-forecast, worst-case, Newsvendor and expected utility maximization (EUM), are employed to facilitate decision-making for market participants on the DAM using the developed CP methods to enhance reliability. The purpose of those bidding strategies is to determine the optimal quantity a PV power supplier should offer in the DAM to yield a high overall profit with minimal energy imbalance. Lastly, the performance of this framework is evaluated using actual data from the Netherlands. This research is mainly relevant for market participants to increase profit while being aware of the associated risk, and for grid operators to improve insights into the expected grid loading. The main contribution can be summarized as follows:

\begin{itemize}
    \item Proposing a novel framework using CP to aid decision-making for PV power market participants on the DAM;
    \item Developing, applying, and evaluating various combinations of CP with bidding strategies, using actual weather and energy market data from the Netherlands. 
\end{itemize}

The remainder of the paper is organized as follows. Section~\ref{sec:2} describes the regression, CP and stochastic optimization methods. The input data and simulation outline are described in Section~\ref{sec:3}. The results are then presented in Section~\ref{sec:4}. The discussion is provided in Section~\ref{sec:5}, with pointers for future work. Finally, the paper is concluded in Section~\ref{sec:6}.

\section{Methods}
\label{sec:2}

Classical machine learning models fail to properly estimate the uncertainty of their predictions \cite{MLUncertainty}, and most quantile regression methods have underlying assumptions of the data distribution, which might not reflect reality. This is where CP comes into play. CP is a relatively new methodology, witnessing a growing body of literature on the subject annually. In Scopus, it has been rising from no publications in 2006 to 30 in 2015 and to 73 in 2022\footnote{Using scope \emph{title, abstract and keyword} on the 6$^{th}$ of February 2023 with the search query: (”conformal predict*” OR ”conformal inference” OR ”conformal regressor”)}. There are conformal regressors and conformal classifiers, but as this study focuses on regression, any mention of CP refers to conformal regressors. CP transforms point predictions into uncertainty intervals without the need for distributional assumptions on the data \cite{GentleIntro}. The uncertainty intervals are rigorous, indicating a probabilistic guarantee of covering the true outcome. In other words, CP guarantees marginal coverage, for which the user chooses an error rate, $\alpha$.

\begin{figure}[!ht]
\centering
\includegraphics[width=3.5in]{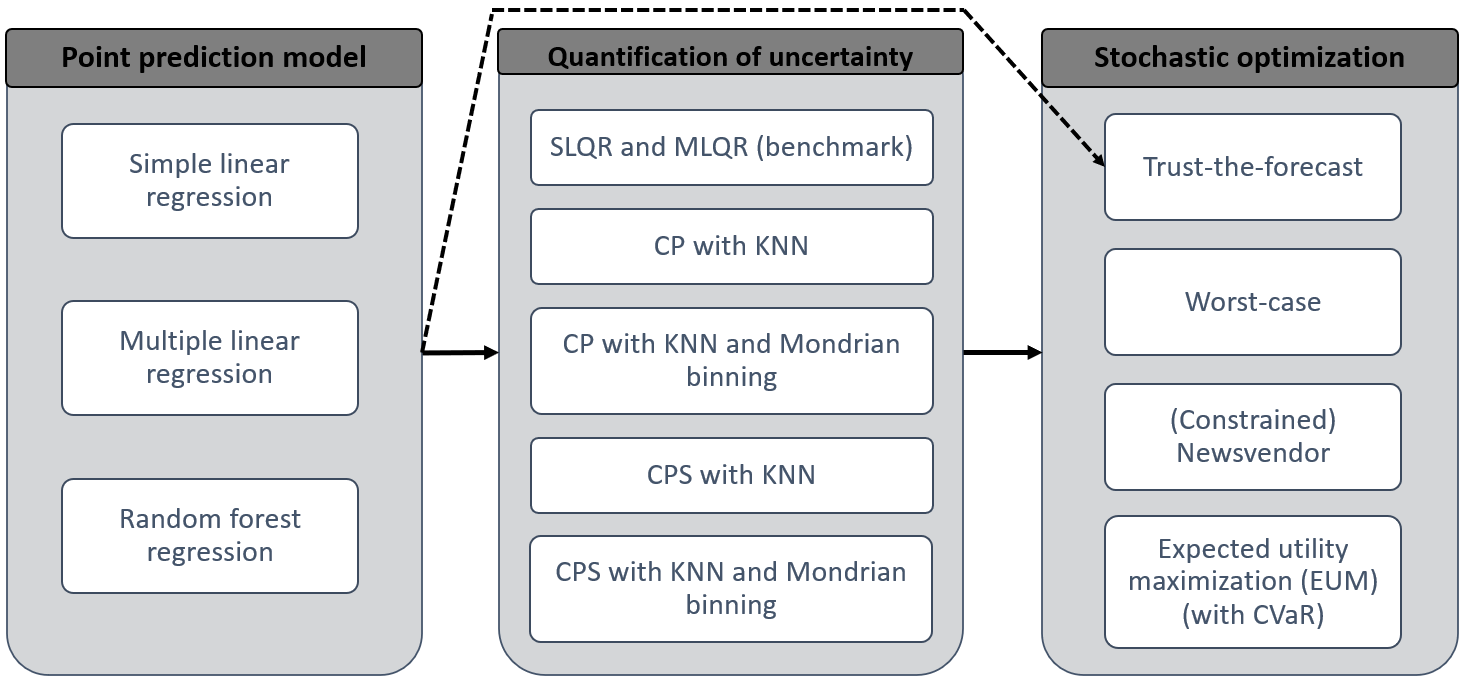}
\caption{Diagram of the proposed framework with the prediction methods and decision-making strategies used.}
\label{fig:PracFlow}
\end{figure}

Fig.~\ref{fig:PracFlow} identifies three steps in the proposed framework to determine DAM bids for PV generation using a CP method. In the first step, point predictions are made for PV generation, after which the uncertainty of these predictions is quantified. Lastly, the optimal bids for the DAM are determined. The following sections introduce the models used in each step.

\subsection{Point prediction models}
\label{subsec:MethodsSQ1}
This study considers three models for making PV power point predictions: Simple and multiple linear regression (SLR \& MLR, respectively) and random forest regression (RFR). Linear and tree-based models are selected, because those have shown to perform at least as good as more complicated models for the forecasting of electricity consumption and PV power \cite{DLvsML1, VISSER2022267}. The superiority of tree-based models on tabular data stems from their capacity to accommodate irregular patterns and maintain robustness  \cite{DLvsML2}. RFR is particularly adopted since it is the best-performing day-ahead PV power forecast model \cite{VISSER2022267}, and it typically yields more efficient conformal predictors compared to other models \cite{LitSQ2-1-1}.

Before making point predictions, the input data should be split into a training, calibration and test set. The training set is used to determine the optimal predictor feature set using forward subset selection with 10-fold cross-validation and for hyperparameter tuning for the RFR model. Eventually, the model is trained using the training dataset with the optimal hyperparameter and feature sets, and the datapoints in the calibration and test datasets are predicted with the fitted model.

\subsection{Uncertainty quantification methods}
For the quantification of uncertainty, different CP methods are considered in this work, which are introduced below and summarized in Table \ref{table:CPMethods}. Simple and multiple linear quantile regression (SLQR \& MLQR, respectively) serve as benchmark models. For these models, the same optimal predictor feature sets as for the SLR and MLR models have been assumed.

\begin{table}[!ht]
\renewcommand{\arraystretch}{1.1}
\centering
\caption{Overview of the considered CP methods in this study.}
\label{table:CPMethods}
\begin{tabular}{p{0.20\linewidth} p{0.70\linewidth}}
\hline\hline
Abbreviation & Model \\
\hline
M1 & Basic CP \\
M2 & CP with KNN \\
M3 & CP with KNN and with Mondrian binning \\
M4 & CPS with KNN  \\
M5 & CPS with KNN and with Mondrian binning \\
\hline\hline
\end{tabular}
\end{table} 

\subsubsection{Basic CP}
\begin{figure}[!ht]
\centering
\includegraphics[width=3.3in,trim={1cm 0cm 1cm 1.5cm},clip]{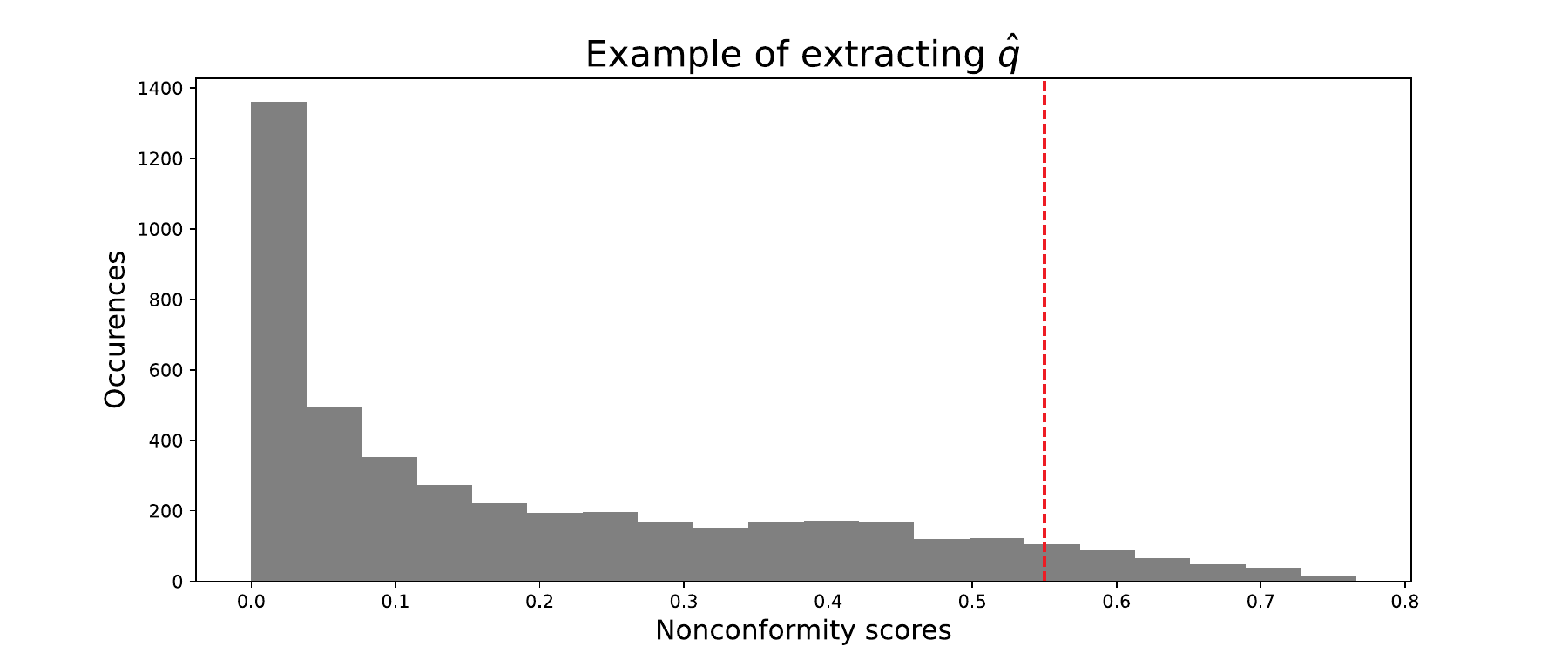}
\caption{Example of how \^{q} is extracted from the sorted nonconformity scores where $\alpha$ is chosen to be 0.10 following that $\hat{q}$ is calculated as the percentile corresponding to $(1-\alpha)$. The red line shows the value for \^{q}, in this case 0.55, where a fraction $\alpha$ of the nonconformity scores exceeds \^{q}.}
\label{fig:ExampleQ}
\end{figure}

The first step for basic CP is to apply a point prediction model to the calibration dataset, after which the residuals of these predictions are extracted. The absolute values of the residuals are used as nonconformity scores, which is a measure indicating how 'atypical' a certain datapoint is. After this, the variable \^{q} is determined based on the chosen value of the error rate $\alpha$ and the sorted nonconformity scores, such that a fraction $\alpha$ of the calibration datapoints has nonconformity scores exceeding \^{q} (see Fig.~\ref{fig:ExampleQ} where $\hat{q}$ is calculated as the percentile corresponding to $(1-\alpha)$). Lastly, the point prediction model is used to predict the points in the test set and \^{q} is both added and subtracted from the point predictions to derive the CP intervals. 

\subsubsection{CP with uncertainty scalars}
The basic CP method guarantees marginal coverage, i.e., the realized values that are outside of the CP interval do not exceed the chosen error rate, $\alpha$. However, it does not account for conditional coverage \cite{GentleIntro}, i.e., the fact that the uncertainty of the predictions differs per datapoint. The property of a method to give wider intervals for points that are harder to predict is called adaptivity. 

The CP method with uncertainty scalars aims to increase adaptivity to approximate conditional coverage. In this method, the difficulty of forecasting a specific datapoint is determined by the uncertainty-estimate, which is the average value of the residuals of the $k$-nearest neighbors (KNN) in the calibration dataset that are most similar with respect to the predictor feature scores. An iterative empirical process in this study has shown that the optimal value for the number of neighbors $k$ is 50. A lower value for the number of neighbors leads to overfitting, while a higher value reduces the adaptivity. The nonconformity scores for a specific datapoint are then equal to the absolute value of the residuals divided by their uncertainty-estimates. Subsequently, the variable \^{q} is determined in the same way as with basic CP, using $\alpha$ and the new nonconformity scores. As a last step, the prediction intervals are determined by adding and subtracting the product of \^{q} and the uncertainty-estimate to the point prediction value.

\subsubsection{CP with Mondrian binning}
CP can also be performed after splitting both the calibration and test datasets into Mondrian categories (i.e., also called Mondrian binning or simply binning). In Mondrian binning, the predicted values for PV power from the calibration dataset are sorted and subdivided into a predefined number of equally-sized bins. Subsequently, the datapoints in the test dataset are assigned to one of those bins based on the value of the point prediction. The CP values are determined in the same way as before, considering only the values of the residuals in the corresponding bin. In previous research, CP with binning was found to outperform basic CP by differing the interval widths between bins and thus creating adaptivity \cite{Mondrian}. Based on empirical evaluation, the optimal number of bins in this study has been found to be 15, as more than 15 bins leads to overfitting, while fewer bins reduce the adaptivity.

\subsubsection{CPS}
\begin{figure}[!ht]
\centering
\includegraphics[width=3.3in,trim={1cm 0.5cm 1cm 1.2cm},clip]{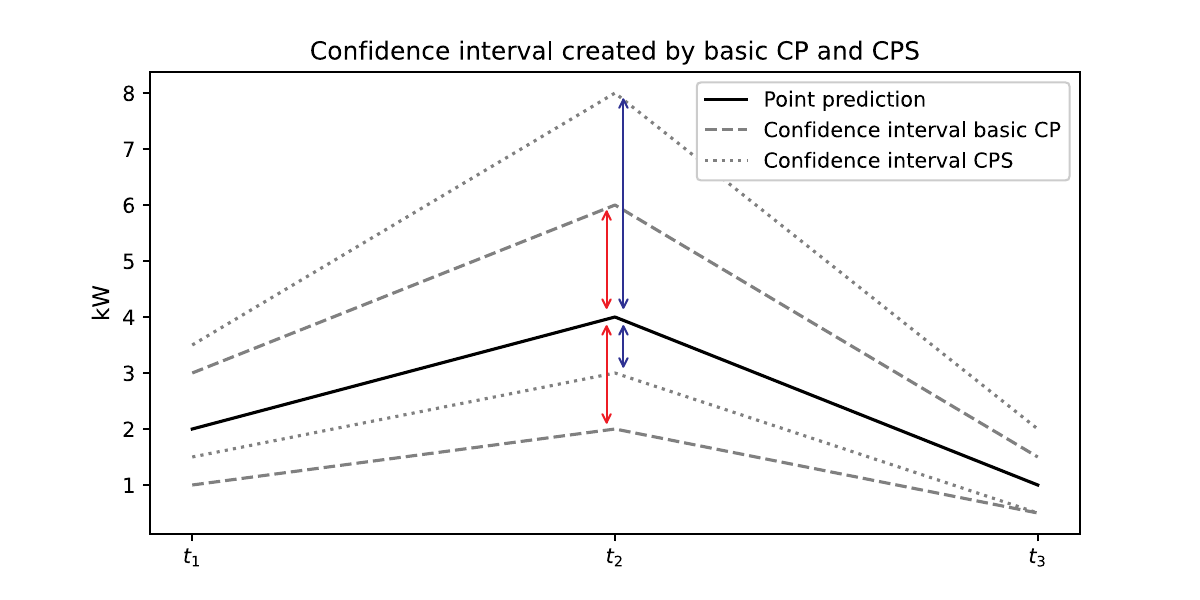}
\caption{Theoretical example showing prediction intervals created by basic CP and CPS showing that the upper and lower sides of the interval are not necessarily equally sized for CPS.}
\label{fig:CPvsCPS}
\end{figure}

An upcoming CP variant is conformal predictive systems (CPS) which outputs conformal predictive distributions (CPD), i.e. CDFs. CPS uses the residuals instead of the absolute values of the residuals as nonconformity scores \cite{PPBoström}. Consequently, a prediction is not necessarily centered in the middle of an interval. In other words, the intervals can be 'shifted' and the left and right hand side of the intervals are not equal by definition (see an example in Fig.~\ref{fig:CPvsCPS}). Therefore, CPS is more flexible than CP and thus preserves more information. Most of the variants for CP can also be applied to CPS. In this study, both KNN and binning are used in combination with CPS.

\subsection{Bidding strategies}
\label{subsec:MethodsSQ3}
Subsequently, different bidding strategies are applied to the test dataset to determine the optimal quantity bids for PV power in the DAM. All strategies are evaluated on both profit and imbalance, assuming perfect foresight of the DAM prices. A perfect strategy yields high revenue with minimal imbalance. The profit for evaluation is based on the proposed bids times the DAM price, and on the imbalance volumes multiplied by their respective actual RTM prices. The imbalance volumes are based on the difference between the bids and the actual PV power. Five bidding strategies are considered in this work.

\subsubsection{Trust-the-forecast} This strategy does not consider the uncertainty in predictions and is applied to the point prediction methods (SLR, MLR and RFR), the quantile prediction methods (SLQR and MLQR) and the CP methods. For the point prediction models, the bid size is equal to the point prediction. For the quantile prediction methods and the CP methods, the bid size is equal to the median value of the uncertainty range.

\subsubsection{Worst-case} For the quantile prediction methods, the DAM bids in this strategy are equal to the predicted 0.05 quantiles. Similarly, for the CP methods, the bid is equal to the lower bound of the prediction interval with $\alpha$=0.10.

\subsubsection{Newsvendor} This bidding strategy is based on the Newsvendor problem \cite{arrow1951optimal}. This is a classical mathematical model in operations management used to determine optimal inventory levels, based on a product's purchase and retail price. This work uses the ratio between the imbalance price deltas to determine the optimal probability quantile; if the average upward imbalance price delta (i.e., the penalty for generating less than the DAM bid) is higher than the average downward imbalance price delta (i.e., the penalty for generating more than the DAM bid), it is generally beneficial to bid higher than the point prediction. As further explained in Section~\ref{sec:3b}, this study uses imbalance price clusters as scenarios for optimization. Eq.~\ref{eq:NV} is applied to each cluster to determine the optimal probability quantile (\( NV_{pq}\)) of the corresponding cluster. 
\begin{align}
 \label{eq:NV}
\text{NV}_{pq} = \frac{\Delta_{\text{down}}}{\Delta_{\text{down}}+\Delta_{\text{up}}} = \frac{p_{\text{DAM}}-p_{\text{down}}}{(p_{\text{DAM}}-p_{\text{down}})+(p_{\text{up}}-p_{\text{DAM}})}
\end{align}

The DAM bid in this strategy is equal to the predicted PV value corresponding to the weighted average $NV_{pq}$ for all imbalance price clusters. Section~\ref{sec:3b} explains how quantile values are determined for CP models. As proposed by \cite{zugno}, this approach can be extended by adding constraints in probability and decision space. When constraining the probability space, the $NV_{pq}$ values for an imbalance price cluster that fall outside a predefined quantile interval range are updated to the closest quantile value within this range. With decision space constraints, the bid volume can only differ by a predefined percentage from the point prediction.

\subsubsection{Expected utility maximization (EUM)} This strategy aims to maximize the expected profit, based on RTM price and PV power scenarios. PV scenarios are generated using the prediction intervals of the forecasts. RTM price scenarios are generated using historical RTM prices, as further detailed in Section \ref{sec:3b}. The detailed mathematical problem formulation for the EUM strategy is outlined in Appendix \ref{subsec:PF1}.

In addition to the basic EUM strategy, this work also considers the EUM strategy in combination with the conditional value at risk (CVaR), which is a risk metric for the tail risk of the model's objective function \cite{Rockafellar2000}. Depending on the considered confidence level $\gamma$, the CVaR is equal to the average profit of the 1-$\gamma$ scenarios with the lowest profit. In this model, both the expected costs and the CVaR are considered in the objective function, where a specific weight $\beta$ is given to the CVaR in the objective function. The full problem formulation for the EUM with the CVaR is shown in Appendix \ref{subsec:PF3}.

\subsubsection{Perfect information} In this reference case, the bids are equal to the actual PV power output and, therefore, the imbalance volume is zero. 

\section{Data inputs and simulation outline}
\label{sec:3}

\subsection{Forecasting setup}
\label{sec:3a}
The different methodological steps outlined in Section~\ref{sec:2} are applied to an open-source dataset with power measurements of 175 PV systems in the province of Utrecht, the Netherlands \cite{Data175PV}. The PV power measurements have a one-minute resolution and cover January 2014 until December 2017. The values are normalized to kW per kWp and are converted to the average hourly values. Subsequently, the aggregated power for all 175 PV systems is determined. In the data pre-processing step, night values are set to zero and outliers are removed.

\begin{figure}[!ht]
\centering
\includegraphics[width=3.8in,trim={1.3cm 0.5cm 0.5cm 1cm},clip]{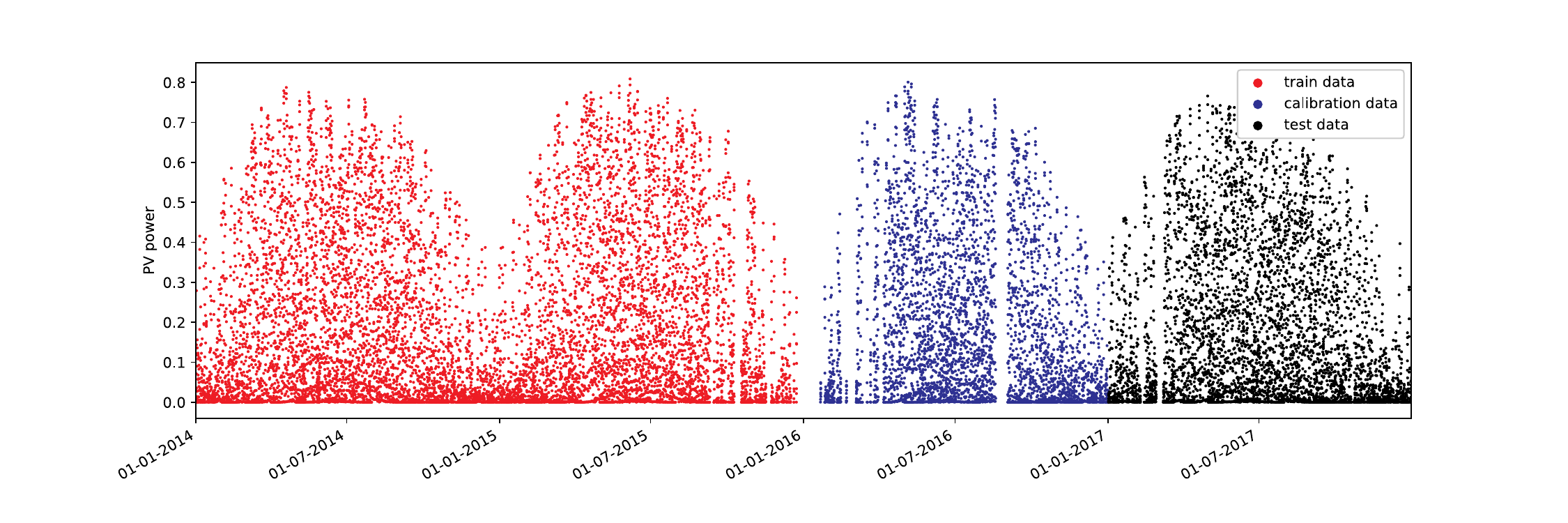}
\caption{The normalized PV power values over time split into train, calibration and test data. The 1$^{st}$ of January of 2016 and 2017 are used as breaking points to split the dataset.}
\label{fig:DataSplit}
\end{figure}

The data is split by date with years 2014 and 2015 as training dataset, 2016 as calibration dataset and 2017 as test dataset as shown in Fig.~\ref{fig:DataSplit}. The data covers around four to five thousand datapoints in each year for sunrise hours. Night values are always zero and therefore the model is trained, calibrated and tested using the day values only. 

Different predictor variables are considered in the forecasts. First, temporal and physical properties are considered, including the hour of the day (presented as \(cos(\frac{2\pi*HoD}{24})\) and \(sin(\frac{2\pi*HoD}{24})\) with HoD as the hour of the day to reflect its cyclic nature \cite{HoD}). Besides, clear sky irradiance, solar zenith and solar azimuth angle are used, which are all determined using the Python PVlib package \cite{holmgren2018pvlib}. In addition, weather forecast data from the Meteorological Archival and Retrieval System of the European Centre for Medium-Range Weather Forecasts (ECMWF)  \cite{ecmwf} has been used as predictor variables. This dataset contains hourly weather predictions for January 2014 till December 2017 on noon of day T for day T+1, matching the requirements of day-ahead solar PV forecasting (e.g., 24 hours ahead with a one-hour resolution). It contains predictions on variables such as surface pressure, cloud cover, wind speed, temperature, precipitation and solar irradiance. The predictions are for \emph{De Bilt} in the province of Utrecht. 

After the forward subset selection, the optimal predictor variable set for the MLR model is found to be the hour of the day, zonal wind speed, total cloud cover, surface solar radiation (SSR) and \(cos(\frac{2\pi*HoD}{24})\). For SLR, the considered predictor variable is the day-ahead forecast of the surface solar radiation downwards (SSRD), as it has the highest correlation with PV power. For the SLQR and MLQR models, the same optimal predictor variable sets as the SLR and MLR models, respectively, have been assumed. The hyperparameter tuning for RFR indicates that the model performs best with 375 trees and with a maximum of 3 features per tree.

\subsection{Optimization setup}
\label{sec:3b}
For the different strategies used to determine the quantity bid for the DAM, DAM prices for the Netherlands in the considered time period have been extracted from the ENTSO-E Transparency Platform \cite{entso-e}. The up- and down-regulation prices in the RTM come from TenneT \cite{tennetData}, the Dutch TSO. 

All CP methods, except the basic CP, are performed with the help of the crepes package (version 0.1.0) \cite{CPpackages} in Python. The Newsvendor strategy is considered without any constraints and with 10\% and 20\% constraints in both the decision and probability space. Quantile values for the CP models are determined by taking the upper or lower bounds of uncertainty intervals after running the model for different values of $\alpha$.

\begin{figure}[!ht]
\centering
\includegraphics[width=3.5in,trim={1cm 0.5cm 0.7cm 1.8cm},clip]{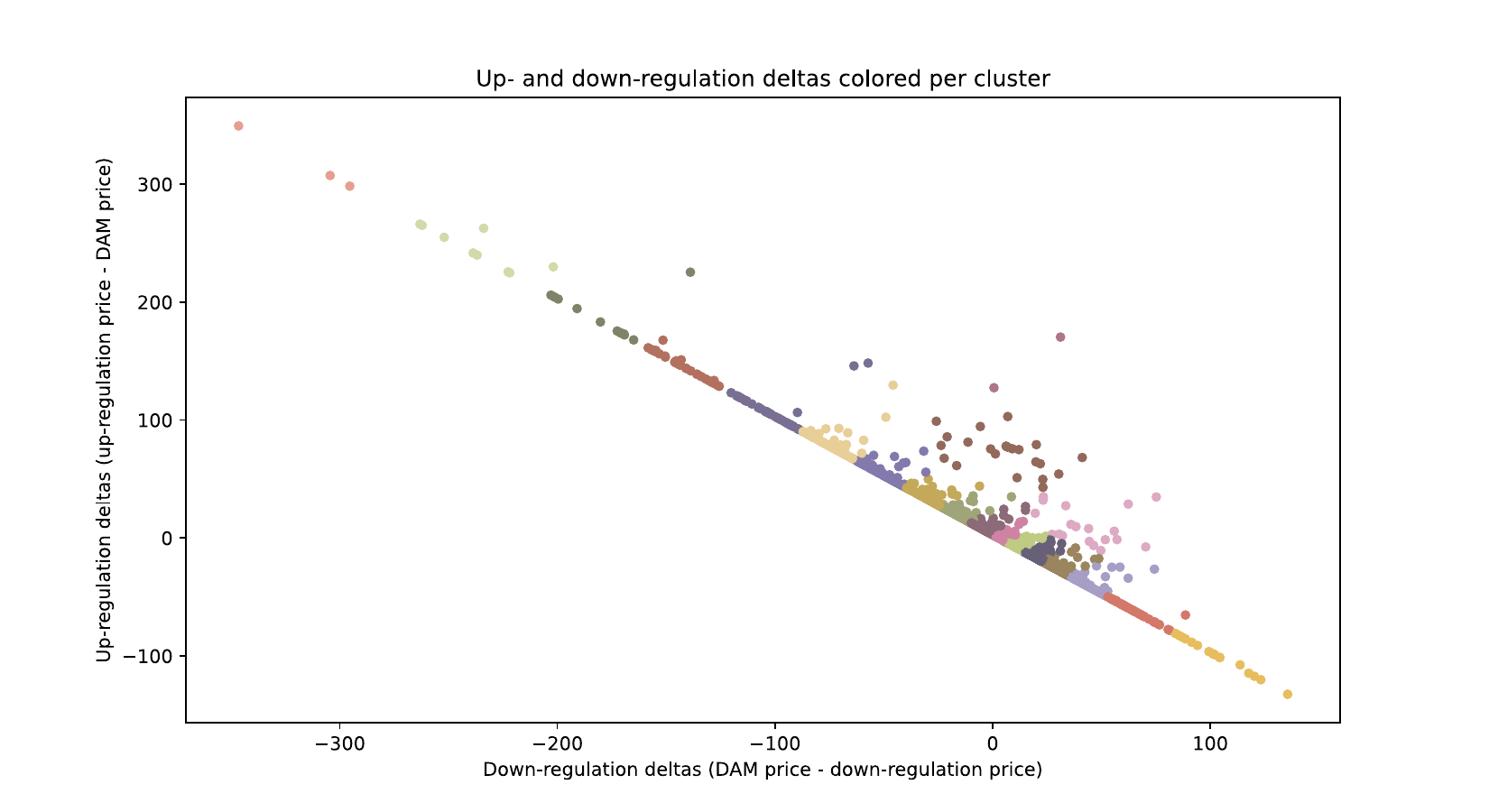}
\caption{The up-regulation deltas against the down-regulation deltas in the calibration dataset colored by their clusters.}
\label{fig:Kmeans}
\end{figure}

The EUM models are run using Gurobi (version 10.0.1) as a solver in Python 
(version 3.8.10). For every timestep, 99 PV power scenarios were considered which equal the 0.01 to 0.99 quantile predictions. Since the RTM prices are highly dependent on the DAM prices, historical imbalance price deltas (i.e., the difference between the DAM price and the RTM price) are used instead of absolute RTM prices. Hourly average RTM prices were used to generate imbalance price deltas. This study has taken all imbalance price deltas in the calibration period and clustered them into twenty clusters using k-means clustering.  Subsequently, the average imbalance price delta for each cluster is taken, and added to the DAM price in the test dataset to get twenty RTM price scenarios per timestep. Fig.~\ref{fig:Kmeans} shows the clusters for the up- and down-regulation deltas. The up- and down-regulation deltas are related, so each scenario represents a specific combination of up- and down-regulation deltas. In the optimization models, the sizes of the clusters are considered as weights for the scenarios.

\section{Results}
\label{sec:4}
\subsection{Point prediction models}
\label{sec:4a}
Table \ref{table:SQ1error} shows the adjusted R$^{2}$ for each point prediction model on the test dataset. The RFR performs best, followed by MLR and SLR, in that order.

\begin{table}[!ht]
\renewcommand{\arraystretch}{1.3}
\centering
\caption{Model performance of the point prediction models}
\label{table:SQ1error}
\begin{tabular}{p{0.375\linewidth} p{0.375\linewidth}}
\hline\hline
Model & Adjusted R$^{2}$ \\
\hline
SLR & 0.806\\
MLR & 0.821\\
RFR & \textbf{0.854}\\
\hline\hline
\end{tabular}
\end{table} 

\subsection{Uncertainty quantification methods}
\label{sec:4b}
Since CP methods do not make use of quantile forecasts, standard quantile forecasting performance metrics such as the continuous ranked probability score (CRPS) cannot be used to evaluate the forecasting performance of the CP methods. Instead, this section assesses the performance of CP methods using the weighted interval score (WIS), which approximates the CRPS \cite{WIS}. This metric is based on the interval score (IS), which considers two elements for a specific value of $\alpha$, namely: sharpness and calibration. Sharpness is the average interval width between the lower and upper bound of the CP confidence interval for a specific value of $\alpha$, and calibration is the sum of the penalties for datapoints in the test dataset outside of the interval. These penalties become more severe for lower error rates (i.e., decreasing values of $\alpha$). The WIS can be determined by combining the IS for multiple error rates \cite{WIS}. This work considered values of $\alpha$ between 0.02 and 0.98 with steps of 0.02 when determining the WIS for CP models. For the linear quantile regression (LQR) models, the WIS is determined by considering the two different quantile values as the lower and upper bound as the confidence intervals. For instance with $\alpha=0.12$, the 0.06 and 0.94 quantile prediction values formed the lower and upper bound of the confidence interval. Lower WIS values indicate a better-performing CP model. The reader may refer to \cite{WIS} for more elaboration on the IS and WIS metrics.

\begin{table}[!ht]
\renewcommand{\arraystretch}{1.3}
\centering
\caption{CP evaluation metrics for the SLQR and the different CP methods using RFR as a point prediction model.}
\label{table:ResultsSQ2}
    \centering
    \begin{tabular}{|c|c|c|c|c|}
    \hline
        Method & Version & WIS  & Sharpness & Calibration \\ \hline
        LQR & SLQR & 0.168 & 0.061 & 0.106 \\ \hline
        M1 & RFR & 0.152 & 0.040 & 0.111 \\ \hline
        M2 & RFR & 0.142 & 0.044 & 0.098 \\ \hline
        M3 & RFR & \textbf{0.140} & 0.061 & 0.079 \\ \hline
        M4 & RFR & 0.143 & 0.045 & 0.098 \\ \hline
        M5 & RFR & 0.142 & 0.057 & 0.085 \\ \hline
    \end{tabular}
\end{table}

Table \ref{table:ResultsSQ2} presents the WIS scores for the different considered CP methods and for SLQR. The RFR model is used as the point prediction model, due to its superior forecasting performance compared to SLR and MLR. Different aspects can be concluded from Table \ref{table:ResultsSQ2}. First, the forecasting performance of all considered CP models is higher than the reference case (SLQR). Second, the more-advanced CP models (M2-M5) outperform the basic CP method (M1). Among the more-advanced CP models, the models that consider CP (M2 \& M3) perform better than the models that use CPS (M4 \& M5). Similarly, the use of Mondrian binning (M3 \& M5) results in better WIS values.  

\begin{figure}[!ht]
\centering
\includegraphics[width=2.5in,trim={0.6cm 0cm 0cm 0cm},clip]{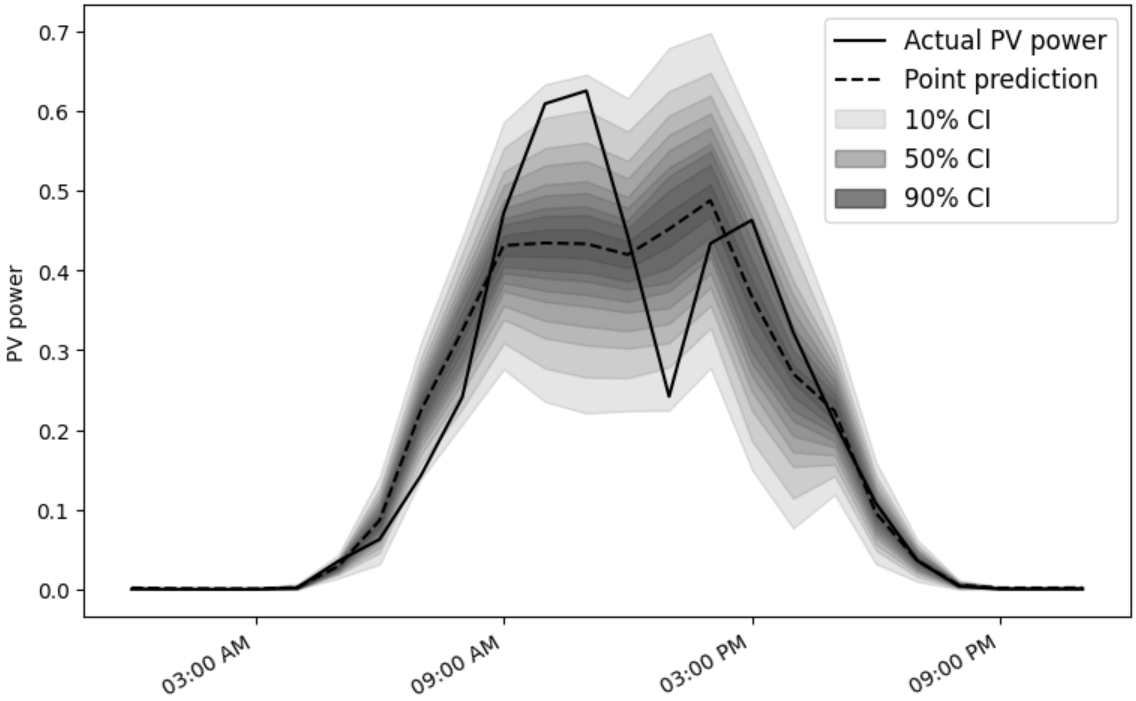}
\caption{Example (June 5th 2017) of normalized point predictions by the RFR model and confidence intervals from M3 after RFR.}
\label{fig:ExampleCP}
\end{figure}

Table \ref{table:ResultsSQ2} shows that the best-performing models in terms of WIS score well on calibration (i.e., lower number of realized values outside the uncertainty interval), but show a lower sharpness, indicating that the uncertainty intervals are larger. Overall, it can be concluded that using RFR combined with CP with k-nearest neighbors and binning (M3) yields the best-weighted interval score. Fig.~\ref{fig:ExampleCP} shows an example of the confidence intervals predicted by this method for one day. 

\subsection{Bidding strategies}
Table \ref{table:SQ3errorsum} provides an overview of the realized profit and imbalance volumes for the different bidding strategies. It reports the results for the best-performing uncertainty models. 

As discussed in Section~\ref{subsec:MethodsSQ3}, the benchmark model is the ideal case with perfect information regarding the PV generation output. As expected, the profit in this model is highest and imbalance volumes are by definition 0. The maximum profit using the trust-the-forecast strategy is equal to 92.4\% of the profit with perfect information. The 'worst-case' bidding strategy yields the lowest profit and highest imbalance volumes. This is mostly caused by the conservative nature of the PV bids with this strategy, which induces high surpluses (i.e., a realized PV power that is higher than the DAM bid), leading to imbalance costs in the RTM. The profit with the Newsvendor strategy is similar to the profit with the 'trust-the-forecast' strategy, although imbalance volumes are slightly higher. Adding the constraints in decision and probability space for this strategy results in a higher profit. Similarly, the profit for the EUM strategy without constraints is slightly lower (0.4\%) than the profit with the 'trust-the-forecast' strategy. The highest profit can be achieved when using the EUM strategy with CVaR. When considering the CVaR, the bids are more conservative and thus lower, leading to fewer deficits and more surpluses, increasing the imbalance volumes. The actual RTM prices in the test dataset make it beneficial to have a surplus, resulting in higher profit.

\begin{table}[!ht]
\renewcommand{\arraystretch}{1.1}
\centering
\caption{Realized profit and imbalance volumes for the different considered bidding strategies. For every bidding strategy the results are shown for the best-performing combination of a point prediction model and a CP method.}
\label{table:SQ3errorsum}
\begin{tabular}{p{0.49\linewidth} p{0.14\linewidth} p{0.08\linewidth} p{0.12\linewidth} }
\hline\hline
Bidding strategy & CP method & Profit [€] & Imbalance [\%] \\
\hline
\textit{Benchmark: Perfect information} & ~ & \textit{34422} & \textit{0.00}\\
Trust-the-forecast & RFR-M5 & 31821 & 4.98 \\
Worst-case & RFR-M4 & 30648 & 8.81\\
Newsvendor (no constraint) & RFR-M5 & 31815 & 5.23 \\
\textbf{Newsvendor (10\% in decision space)} & \textbf{RFR-M5} & \textbf{31863} & \textbf{5.10} \\
Newsvendor (10\% in probability space) & RFR-M5 & 31828 & 5.05 \\
EUM (no constraint) & RFR-M5 & 31683 & 5.07 \\
EUM (CVaR with $\gamma$=0.75 \& $\beta$=0.1) & RFR-M5 & 31852 & 5.60 \\
\textbf{EUM (CVaR with $\gamma$=0.6 \& $\beta$=0.1)} & \textbf{RFR-M5} & \textbf{32033} & \textbf{5.54} \\
EUM (CVaR with $\gamma$=0.6 \& $\beta$=0.2) & RFR-M5 & 32761 & 6.76 \\
\hline\hline
\end{tabular}
\end{table}

Table \ref{table:profitpercentage} compares the profits between the different uncertainty models for the different bidding strategies. For the EUM strategies the LQR models are outperformed by the CPS models, indicating that using CPS to account for uncertainty is superior to using quantile predictions. In general, the models that consider CPS (M4 and M5) perform better than the CP models without CPS (M2 and M3). Also, the CP models that consider Mondrian binning (M3 and M5) generally outperform their counterparts without Mondrian binning (M2 and M4). 

The results show that the CPS methods yield a higher profit than the CP methods for the Newsvendor and the EUM strategies, despite having a higher WIS than the CP methods. Although the WIS is worse for CPS, the additional flexibility of CPS improves the stochastic decision-making. This flexibility arises from the fact that the CPS point predictions are not necessarily centered in the middle of the prediction intervals (see Fig.~\ref{fig:CPvsCPS}).

\begin{table}[]
\centering
\caption{Realized profit (as a \% of max. profit for a specific strategy) for different uncertainty models. With 'con' standing for 'constraint' and 'dec' for 'decision space', and with $\gamma$=0.6 \& $\beta$=0.1 for the CVaR.}
\label{table:profitpercentage}
\begin{tabular}{lcccc}
\hline \hline
              & \textbf{\begin{tabular}[c]{@{}c@{}}Newsvendor\\ (no con.)\end{tabular}} & \textbf{\begin{tabular}[c]{@{}c@{}}Newsvendor\\ (10\% dec.)\end{tabular}} & \textbf{\begin{tabular}[c]{@{}c@{}}EUM\\ (no con.)\end{tabular}} & \textbf{\begin{tabular}[c]{@{}c@{}}EUM\\ (with CVaR)\end{tabular}}                              \\ \hline
\textbf{SLQR} & 99.67\% & 99.36\% & 98.64\% & 98.48\% \\
\textbf{MLQR} & 99.46\% & 99.35\% & 98.96\% & 98.88\% \\
\textbf{RFR-M2} & 98.90\% & 98.90\% & 99.26\% & 99.06\% \\
\textbf{RFR-M3} & 99.16\% & 98.99\% & 98.74\% & 98.95\% \\
\textbf{RFR-M4} & 98.90\% & 98.92\% & 99.27\% & 99.09\% \\
\textbf{RFR-M5} & \textbf{100.00\%} & \textbf{100.00\%} & \textbf{100.00\%} & \textbf{100.00\%} \\ \hline \hline
\end{tabular}
\end{table}

\section{Discussion}
\label{sec:5}
This section reflects on the different methodological assumptions made in this work, alongside with the limitations associated with these assumptions. In addition, the implications of this work is discussed.

Due to the extensive scope of this study, several simplifications were necessary. For instance, it was assumed that the DAM prices are perfectly known and that the RTM price clusters accurately represent the actual RTM prices. In addition, we limited our analysis to linear and tree-based point prediction models, and considered a limited number of RTM price clusters, PV scenarios and constraint percentages for the Newsvendor strategy. Besides, the considered number of neighbors with KNN and the number of bins with Mondrian binning was determined through a simplified empirical evaluation. The feature selection step considered the same optimal feature sets for the SLR and MLR models. Despite these simplifications, we are confident that the overall trends are accurately portrayed. Future research could delve into more complex models and examine the impact of these simplifications. Additionally, future research could include the intraday market, which offers market participants the opportunity to correct forecasting errors, enhancing flexibility and potentially leading to different bidding strategies.

When interpreting the results, one should keep in mind that this study considered different aggregation steps which reduce the number of extreme values, for instance by averaging the 15-minute RTM prices to hourly values and by aggregating the 175 PV power systems. This mainly impacts the CVaR which restricts the extreme losses and, therefore, mostly functions well for extreme scenarios.

It should also be considered that the selection of the preferred method can differ between energy suppliers. Risk aversion leads to different considerations with respect to profit and imbalance. When mostly considering profit, RFR-M5 combined with the EUM with CVaR ($\gamma$=0.6 \& $\beta$=0.1) can be chosen, which yields 93\% of the potential profit with minimal imbalance. However, if a low imbalance is more important, the Newsvendor strategy with a 10\% constraint in the decision space is more fitting.

Lastly, this research is limited to quantity bids and does not include a proposed price bid. Before this framework can be implemented, it should be extended with a method to determine the price bids. 

\section{Conclusions and Future Work}
\label{sec:6} 
This paper proposes a novel framework leveraging CP, an emerging probabilistic forecasting method, to enhance decision-making in electricity markets. In the first step, day-ahead point predictions of PV power are generated. Subsequently, several LQR and CP methods are used to quantify the uncertainty of the point predictions. Lastly, the prediction intervals and CPD are used as input for several decision-making strategies to determine DAM quantity bid under uncertainty. One of the main conclusions is that CP methods outperform other methods that quantify uncertainty, such as linear quantile regression. In addition, results show that CPS with KNN and binning performs best on both profit and imbalance for almost all considered market bidding strategies. When mainly aiming for high profits, the EUM strategy with CVaR ($\gamma$=0.6 \& $\beta$=0.1) is recommended. When valuing a lower imbalance more highly, the Newsvendor strategy with a 10\% constraint in the decision space should be used. To further improve the decision-making strategies, intraday markets can be included and the EUM strategy with CVaR should be further explored.

\appendices

\section{Problem formulation EUM strategy}
\label{sec:PF}

\subsection{EUM without constraints}
\label{subsec:PF1}
\begin{equation}
\small
\begin{aligned}
& \underset{p}{\text{maximize}}
& & \sum_{t=1}^{T}\text{DA}_t p_t\! +\!\sum_{c=1}^{C}\sum_{t=1}^{T}\frac{- \text{d}_t \text{w}_{c}(\text{DA}_t\!+\!\text{U}_{c})\! +\! \text{s}_t \text{w}_{c} (\text{DA}_t\!-\!\text{D}_{c})}{N T_{cw}}\\
& \text{subject to}
& & \text{s}_t = \sum_{n=1}^{N} \text{max}(\text{pred}_{n,t}\! -\! p_t, 0), \qquad \forall T \\
&&& \text{d}_t = \sum_{n=1}^{N} \text{max}(p_t\! -\!\text{pred}_{n,t}, 0), \qquad \forall T \\
&&& p_t \in [0,1], \qquad \forall  T
\end{aligned}
\end{equation}
where the indices $t$ and $n$ represent the timestep and PV scenario, respectively, the decision variable $p_t$ (between 0 and 1 due to normalization) represents the bid size, $\text{pred}_{n,t}$ is the prediction of PV power, $\text{DA}_t$ is the DAM price, $\text{U}_{c}$ is the up-regulation price in RTM cluster $c$, $\text{D}_{c}$ is the down-regulation price in the RTM cluster $c$, $\text{d}_t$ is the sum of deficits over all PV-scenarios, $\text{s}_t$ is the sum of surpluses over all PV-scenarios, $\text{w}_{c}$ is a weight for RTM cluster $c$, T is the number of timestamps (i.e., 4514 for the assessment period in this study), C is the number of RTM clusters (i.e., 20 in this study), N is the number of PV-scenarios (i.e., 99 in this study) and $T_{cw}$ is the sum of all cluster weights (i.e., 3778 in this study, which is the number of points in the calibration set).

\subsection{EUM with CVaR}
\label{subsec:PF3}

\begin{equation}
\small
\begin{aligned}
& \underset{p}{\text{maximize}}
& & (1-\beta)\sum_{c=1}^{C}\sum_{n=1}^{N}\sum_{t=1}^{T} \text{pr}_{c,n,t}\! +\! \beta \text{CVaR}\\
& \text{subject to}
& & \text{spv}_{n,t} = \text{max}(\text{pred}_{n,t}\! -\! p_t, 0), \qquad \forall N,T \\
&&& \text{s}_t = \sum_{n=1}^{N} \text{spv}_{n,t} \qquad \forall T \\
&&& \text{dpv}_{n,t} = \text{max}(p_t\! -\!\text{pred}_{n,t}, 0), \qquad \forall N,T \\
&&& \text{d}_t = \sum_{n=1}^{N} \text{dpv}_{n,t}, \qquad \forall T \\
&&& \text{pr}_{c,n,t} = p_t\text{DA}_t\! -\!\text{dpv}_{n,t} (\text{DA}_t\!+\!\text{U}_{c}) +\\
& &
&\ \ \ \ \ \text{spv}_{n,t} (\text{DA}_t\!-\!\text{D}_{c}), \quad \forall  C, N, T \\
&&& \sigma_{c,n,t} \geq \text{VaR} - \text{pr}_{c,n,t}, \quad \forall  C, N, T \\
&&& \sigma_{c,n,t} \leq 0, \quad \forall  C, N, T \\
&&& \text{CVaR}=\text{VaR}\!-\!\frac{1}{1-\gamma}\sum_{c=1}^{C}\sum_{n=1}^{N}\sum_{t=1}^{T}\pi_{c,n}\sigma_{c,n,t} \\
&&& p_t \in [0,1], \qquad \forall  T \\
\end{aligned}
\end{equation}
where $\beta$ is the weight of the CVaR profit in the objective function, $\gamma$ is the confidence level, $\text{dpv}_{n,t}$ and $\text{spv}_{n,t}$ are the deficit and surplus for PV-scenario $n$ at timestamp $t$, respectively, $\text{pr}_{c,n,t}$ is the profit for RTM cluster $c$, PV-scenario $n$ at timestamp $t$. $\sigma$ is an auxiliary variable, $\pi_{c,n}$ is the probability of the scenario and \text{VaR} is the value-at-risk profit. All other variables are defined in Section \ref{subsec:PF1}.


\end{document}